\documentclass{article}

\usepackage[top=1in,bottom=1in,left=1in,right=1in]{geometry}

\usepackage{amssymb,amsthm,amsmath,amscd,color}
\usepackage{epsfig}
\usepackage{subfigure}
\usepackage{enumerate}
\usepackage{hyperref}
\definecolor{darkred}{RGB}{150,0,0}
\definecolor{darkgreen}{RGB}{0,150,0}
\definecolor{darkblue}{RGB}{0,0,200}
\hypersetup{colorlinks=true, linkcolor=darkred, citecolor=darkgreen, urlcolor=darkblue}

\newtheorem{theorem}{Theorem}
\newtheorem{lemma}{Lemma}
\newtheorem{proposition}{Proposition}
\newtheorem{corollary}[theorem]{Corollary}

\theoremstyle{definition}

\theoremstyle{definition}

\DeclareMathOperator{\dist}{dist}

\DeclareMathOperator{\diam}{diam}
\DeclareMathOperator{\diag}{diag}

\newcommand{\reals}{\mathbb R}
\newcommand{\bbR}{\mathbb R}

\newcommand{\be}{\begin{equation}}
\newcommand{\ee}{\end{equation}}

\newcommand{\pr}[1]{P\left( #1 \right)}
\newcommand{\expect}[1]{E\left( #1 \right)}

\def\ba{\mathbf{a}}
\def\bb{\mathbf{b}}

\def\be{\mathbf{e}}

\def\br{\mathbf{r}}
\def\bu{\mathbf{u}}
\def\bv{\mathbf{v}}
\def\bx{\mathbf{x}}
\def\by{\mathbf{y}}
\def\bz{\mathbf{z}}

\def\bB{\mathbf{B}}
\def\bD{\mathbf{D}}

\def\bR{\mathbf{R}}
\def\bU{\mathbf{U}}
\def\bV{\mathbf{V}}
\def\bW{\mathbf{W}}
\def\bZ{\mathbf{Z}}

\def\brD{\mathbf{\mathring{D}}}

\def\brU{\mathbf{\mathring{U}}}
\def\brV{\mathbf{\mathring{V}}}
\def\brW{\mathbf{\mathring{W}}}
\def\brZ{\mathbf{\mathring{Z}}}

\def\bru{\mathbf{\mathring{u}}}
\def\brv{\mathbf{\mathring{v}}}

\def\rD{\mathring{D}}
\def\rL{\mathring{L}}

\def\rW{\mathring{W}}
\def\rZ{\mathring{Z}}

\def\cS{\mathcal{S}}
\def\cX{\mathcal{X}}

\def\eps{\epsilon}
\def\vol{{\rm vol}}

\usepackage{float}
\floatstyle{ruled}
\newfloat{algo}{thp}{lop}
\floatname{algo}{Algorithm}

\begin{document}
\title{Clustering Based on Pairwise Distances \\ When the Data is of Mixed Dimensions}

\author{Ery~Arias-Castro \footnote{Department of Mathematics, University of California, San Diego, La Jolla, CA 92093-0112, USA; \href{mailto:eariasca@ucsd.edu}{eariasca@ucsd.edu}.}}
\date{University of California, San Diego}

\maketitle

{\small
\noindent {\bf Abstract.}  
In the context of clustering, we consider a generative model in a Euclidean ambient space with clusters of different shapes, dimensions, sizes and densities.  In an asymptotic setting where the number of points becomes large, we obtain theoretical guaranties for a few emblematic methods based on pairwise distances: a simple algorithm based on the extraction of connected components in a neighborhood graph; the spectral clustering method of Ng, Jordan and Weiss; and hierarchical clustering with single linkage.  The methods are shown to enjoy some near-optimal properties in terms of separation between clusters and robustness to outliers.  The local scaling method of Zelnik-Manor and Perona is shown to lead to a near-optimal choice for the scale in the first two methods.  We also provide a lower bound on the spectral gap to consistently choose the correct number of clusters in the spectral method. \\

}

\section{Introduction}

In the context of clustering points in an Euclidean space, traditional methods such as $K$-means or Gaussian mixture models, assume that each cluster is generated by sampling points in the vicinity of a centroid.  The resulting clusters are ellipsoidal, and in particular full-dimensional.  Several papers obtain theoretical results in this setting; see e.g.,~\cite{kmeans,dasgupta1999lmg,VW04,1090312,ach05}, and references therein.
In a number of modern applications, however, the data may contain structures of mixed dimensions.  Even the apparently simple case of affine surfaces is a relevant model for a number of real-life situations~\cite{Ma07}.  Our focus here is a more general framework making minimal assumptions on the underlying clusters.  Note that our framework is inclusive of the classical setting.

\subsection{Mathematical framework}
\label{sec:setting} 

We set the ambient space to be the $D$-dimensional unit hypercube $[0,1]^D$, though our results may generalize to other setting such Riemannian manifolds.  In most of the paper, we assume that $D$ is fixed, and discuss the case where $D$ is large in Section \ref{sec:D-large}.  For a positive integer $d \leq D$ and a constant $\kappa \geq 1$, let $\cS_d(\kappa)$ be the class of measurable, connected sets (surfaces) $S \subset [0,1]^D$ such that 
\begin{equation} \label{eq:S-vol}
\forall \bx \in S: \ \kappa^{-1} \eps^d \leq \vol_d(B(\bx, \eps) \cap S) \leq \kappa \eps^d, \quad \forall \eps \in [0, \diam(S)].
\end{equation}
The condition above not only implies that the surface $S$ has (e.g. Hausdorff) dimension $d$ with finite $d$-volume, it also prevents $S$ from being too narrow in some places.  We also define $\cS_0(\kappa)$ as the set of points in $[0,1]^D$.
We let $\cS(\kappa) = \bigcup_{d=0}^D \cS_d(\kappa)$.  

For readers more familiar with function spaces, note that the class $\cS_d(\kappa)$ contains for example the image $f([0,1]^d)$ of locally bi-Lipschitz functions $f:[0,1]^d \to [0,1]^D$ satisfying:
\begin{equation} \label{eq:taylor1}
M^{-1} \|b - a\| \leq \|f(b) - f(a)\| \leq M \|b - a\|, \ \forall a,b \in [0,1]^d,
\end{equation}
for $M$ small enough.    

For $S \subset [0,1]^D$ and $\tau > 0$, define
$$
B(S, \tau) = \{\bx \in [0,1]^D: \min_{\by \in S} \|\bx - \by\| \leq \tau\}.
$$ 
This is the $\tau$-neighborhood of $S$ in $[0,1]^D$ relative to the Euclidean metric.  Given surfaces, $S_1, \dots, S_K \in \cS(\kappa)$, we generate clusters, $\cX_1, \dots, \cX_K$, by sampling $N_k$ points in $B(S_k,\tau)$, the $\tau$-neighborhood of $S_k$, according to a distribution $\Psi_k$ with density $\psi_k$ with respect to the uniform measure on $B(S_k,\tau)$.  We call $\tau$ the noise level or sampling imprecision.  In the noiseless case, $\tau = 0$, the points are sampled exactly on the surface.  We require that $\kappa^{-1} \leq \psi_k \leq \kappa$, so the cluster is somewhat uniformly sampled.  Our results apply without major change for non-compactly supported sampling distributions with fast-decaying tails, such as Gaussian noise.  The classical setting corresponds to either $d = 0$ (centroids), or $d = D$ with $\tau = 0$ (full-dimensional cluster).  Let $N = \sum_k N_k$ be the total number of data points, which we denote by $\bx_1, \dots, \bx_N$.  For later use, define indices $I_k = \{i = 1, \dots, N: \bx_i \in \cX_k\}$.

We assume the clusters do not intersect, and in fact that the underlying surfaces are well-separated:
\begin{equation} \label{eq:delta}
\dist(S_k, S_\ell) := \min_{\bx \in S_k, \by \in S_\ell} \|\bx - \by\| \geq \delta, \quad \forall k \neq \ell.
\end{equation}
The actual clusters are therefore separated by a distance of at least $\delta - 2 \tau$.

\begin{center}
\noindent {\bf Surface clustering task.}
{\it Given data $\{\bx_1, \dots, \bx_N\}$, recover the clusters $\cX_1, \dots, \cX_K$.}
\end{center}

\subsection{In this paper}

We first consider a simple algorithm based on extracting the connected components of a $\eps$-neighborhood graph built using a compactly supported kernel.  We provide conditions guarantying that the algorithm perfectly recovers the underlying clusters in our theoretical setting; this is done in Section \ref{sec:CC}.  This approach may be seen as a precursor to spectral methods, which extract `soft' connected components based on an eigen-decomposition of the Laplacian of the neighborhood graph.  In Section \ref{sec:NJW}, we consider the method introduced by Ng, Jordan and Weiss~\cite{Ng02}, a standard spectral clustering algorithm.  We show that, in our framework, the spectral method operates under very similar conditions as the method based on connected components.  The last method we consider, in Section \ref{sec:single}, is hierarchical clustering with single linkage, which is in some sense equivalent to the method based on connected components.  Note that hierarchical clustering with average or complete linkage are not suitable in our context which includes elongated clusters.  

It turns out that the first two methods are near-optimal in terms of separation between clusters and robustness to outliers.  In Section \ref{sec:sep}, we show that, under low sampling noise, no method can perfectly separate clusters that are closer together than what the first two methods require by more than a poly-logarithmic factor.  For clusters of dimension one or two, we obtain stronger results, showing that all clustering methods have in fact a non-negligible error rate in that same situation.  In Section \ref{sec:detection}, we address the situation where outliers, points sampled elsewhere in space, may be present in the data, and show that the first two methods, properly modified, are able to accurately cluster within logarithmic factors of the best known detection rates~\cite{CTD,ery-thesis}, even though the task of detection is {\it a priori} much easier than the task of clustering.

In the discussion part of the paper, Section \ref{sec:discussion}, we consider the choice of parameters, that is the scale defining the neighborhood graph and, for the spectral method, the number of eigenvectors to extract.  We show that the local scaling method of Zelnik-Manor and Perona~\cite{Zelnik-Manor04}, with a number of nearest neighbors of order slightly larger than $\log(N)$, leads to a near-optimal choice of scale.  As a consequence, computations may be restricted to small neighborhoods without compromising the clustering performance, so that a nearest-neighbor search becomes the computational bottleneck.  We also provide a bound on the eigengap allowing to consistently estimate the number of clusters.  Finally, we discuss how the results generalize to the case where the ambient dimension is very large or even infinite.  

The various proofs are gathered at the end of the paper, in Section \ref{sec:proofs}, with the proof of auxiliary results gathered in Appendices \ref{proof:NJW-orig} and \ref{sec:aux}.  The careful reader will notice that our results could be made non-asymptotic without much change in the arguments.  However, we chose to favor the statement of simple results with concise proofs.

\subsection{Related work}

Neighborhood graphs defined on a random set of points in Euclidean space are sometimes called {\em random geometric graphs}, and have been of interest in modeling networks.  The book by Penrose~\cite{penrose} is a standard reference.  The main difference in our case is that the support of the sampling density may be (close to) singular with respect to the Lebesgue measure.  
Extracting connected components from a neighborhood graph is a natural idea and has been proposed before; we comment on three publications that are particularly relevant to us~\cite{1519716,pelletier07,RePEc:eee:stapro:v:35:y:1997:i:1:p:33-42}.  Maier, Hein and von Luxburg~\cite{1519716} consider $k$-nearest neighbor type graphs and analyze the performance of the resulting clustering algorithm within a slightly more restrictive model where both the clusters and the sampling densities are smooth, and the degree of imprecision is positive, $\tau > 0$.  Within that framework, the results in that paper are non-asymptotic and more precise than our Theorem \ref{th:CC}.  Their emphasis is on choosing $k$ optimally in terms of maximizing the probability of correctly solving the clustering task and on the effect of using different kinds of graphs.  We comment on their work in more detail in Section \ref{sec:discussion}.  In a similar model, Biau, Cadre and Pelletier~\cite{pelletier07} focus on estimating the correct number of clusters based on counting the number of connected components in a $\eps$-neighborhood graph.  Both~\cite{1519716,pelletier07} consider the case where the space between clusters contains points; we call those points outliers and consider this situation in Section \ref{sec:outliers}.  Brito, Ch‡vez, Quiroz and Yukich~\cite{RePEc:eee:stapro:v:35:y:1997:i:1:p:33-42} consider a model similar to ours with all clusters full-dimensional.  They also use a $k$-nearest neighbor graph and show that, when the separation between clusters remains bounded away from zero, choosing $k$ of order $\log N$ makes the algorithm output the perfect clustering; this is similar to our Proposition \ref{prop:NN}.  They also consider a test of non-uniformity, where the alternative is that of points clustered more closely together as opposed to a cluster hidden in a background of uniform points as we consider in Section \ref{sec:detection}.  However, there are no optimality considerations.  In light of~\cite{1519716,pelletier07,RePEc:eee:stapro:v:35:y:1997:i:1:p:33-42}, our contribution is in considering a slightly more general framework, for which we provide short proofs, and in establishing optimality results in terms of separation between clusters and robustness to outliers.

Spectral clustering methods have been specifically developed to work in the kind of framework we consider here~\cite{survey-kernel-spectral}.  Though very popular, few theoretical results are available on the performance of spectral methods under this type of generative model.  Ng, Jordan and Weiss, in their influential paper~\cite{Ng02}, introduce their method and outline a strategy to analyze it; however, no explicit probabilistic model is considered.  The same comment holds for~\cite{990313}.  In~\cite{vonLuxburg08,pelletier08}, spectral clustering is taken to its empirical process limit as the number of points increases; though this provides insight on what spectral clustering is estimating, there is no result on its performance.  This is similar to the analysis in~\cite{LDS_NIPS_06}.  Other papers, such as~\cite{NIPS2003_AA12}, introduce variations on the spectral method and provide theoretical results on computational aspects, not on clustering performance.  Closer in spirit to the present paper is the work of Chen and Lerman~\cite{spectral_theory}, where the authors analyze a multi-way spectral method specifically designed for the case of affine surfaces.  Our contribution here is in providing theoretical guaranties for spectral clustering methods in a rigorous mathematical framework.   In doing that, we provide a concise proof of the main result in~\cite{Ng02} partly based on information that Andrew Ng shared with the author and the proof of~\cite[Th 4.5]{spectral_theory} by Chen and Lerman.  

To our knowledge, the minimax-type bounds on the separation between clusters obtained in Section \ref{sec:sep} are the first of their kind in the context of clustering under a non-parametric model.  In the classical setting, there is some existing literature, though very scarce; we will comment on a paper of Achlioptas and McSherry~\cite{ach05}.  The literature is of course abundant in the context of estimation~\cite{MR1742500,MR1635414,MR1226450,MR902241} and classification~\cite{MR1725115,MR2241192}.  In our arguments, we use the popular approach consisting in reducing the task to a hypothesis testing problem.

\subsection{Additional notation}

Except for $D, K$ and $\kappa$, the parameters such $\delta, \tau$ and $\eps$ vary with $N$.  This dependence is left implicit.  An event $E_N$ holds with high probability if $\pr{E_N} \to 1$ as $N \to \infty$.  We use standard notation, such as: $a \vee b$ for $\max(a, b)$; $a \wedge b$ for $\min(a, b)$;  $a_N \prec b_N$ for $a_N = O(b_N)$; $a_N \asymp b_N$ for $a_N = O(b_N)$ and $b_N = O(a_N)$; $a_N \ll b_N$ for $a_N = o(b_N)$.  

\section{Some standard clustering methods based on pairwise distances}

We describe some common approaches to clustering, all based on pairwise distances.  Each time, we provide sufficient conditions for the method to output the perfect clustering.  These conditions are seen to be necessary up to multiplicative logarithmic factors.  We will see in later sections what these conditions imply in terms of comparative performance.

The first two methods build a neighborhood graph on the data points using an affinity based on pairwise distances:
\begin{equation}
\label{eq:pair-affinity}
\alpha(\bz_1,\bz_2) = \left\{\begin{array}{ll}
\phi(\|\bz_1 - \bz_2\|/\eps), & \bz_1 \neq \bz_2; \\
0, & \bz_1 = \bz_2.
\end{array} \right.
\end{equation}
We assume the kernel $\phi$ is non-negative, continuous at 0 with $\phi(0) = 1$, non-increasing on $[0, \infty)$ and is fast-decaying, in the sense that $s^q \phi(s) = o(1)$ for any $q > 0$.

\subsection{Clustering based on extracting connected components}
\label{sec:CC}

The first algorithm we introduce, Algorithm \ref{algo:CC}, extracts connected components of the neighborhood graph and therefore requires a compactly supported kernel; let $[0,\omega]$ be the support of $\phi$.

\begin{algo}[ht]
\caption{Pairwise clustering based on extracting connected components}
\centering
\begin{tabular}{p{6in}}
\textbf{Input:}\\
\hspace*{.3in} $\{\bx_1,\bx_2,...,\bx_N\} \subset [0,1]^{D}$: the data set\\
\hspace*{.3in} $\eps$: affinity scale\\

\textbf{Output:}\\ 
\hspace*{.3in} A partition of the data into disjoint clusters\\[.1in]

\textbf{Steps:}\\ 
{\bf 1:} Compute the affinity matrix $\bW \in \reals^{N \times N}$, with $W_{ij} = \alpha(\bx_i,\bx_j)$. \\
{\bf 2:} Extract the connected components of $\bW$. \\
{\bf 3:} Accordingly group the original points into disjoint clusters. \\

\end{tabular}
\label{algo:CC}
\end{algo}


\begin{theorem}
\label{th:CC}
Consider the generative model of Section \ref{sec:setting} with surfaces $S_1, \dots, S_K \in \cS(\kappa)$.  Assume that $\delta - 2 \tau > \omega \eps$, with  
\begin{equation} \label{eq:N-cond0}
\eps \gg \max_{k = 1, \dots, K} \max \left\{\begin{array}{l}
(\tau \vee \diam(S_k)) (\log(N_k)/N_k)^{1/d_k}, \\   
(\tau \vee \diam(S_k))^{d_k/D} \tau^{1 - d_k/D} (\log(N_k)/N_k)^{1/D}
\end{array}\right\}.
\end{equation}
Then, Algorithm \ref{algo:CC} is perfectly accurate with high probability.
\end{theorem}
The proof of Theorem \ref{th:CC} is in Section \ref{proof:CC}.
The condition $\delta - 2 \tau > \omega \eps$ means that distinct clusters are separated by $\omega \eps$ and therefore disjoint in the neighborhood graph.  The term in brackets on the right hand side of (\ref{eq:N-cond0}) is actually the order of magnitude for the maximin distance between points sampled from $B(S_k,\tau)$.

{\it Remark.}
In the classical setting where each $S_k$ is a centroid, i.e. $d_k =0$, the algorithm is accurate when 
$$\eps \gg \tau \max_k (\log(N_k)/N_k)^{1/D}.$$

\subsection{Spectral clustering}
\label{sec:NJW}

When using kernels that are not compactly supported, extracting connected components makes little sense as the neighborhood graph is fully connected.  Instead, spectral methods perform an eigen-decomposition of the graph Laplacian.  The spectral method introduced in~\cite{Ng02} uses the Gaussian kernel $\phi(s) = e^{-s^2/2}$.  Note that kernels of compact support are considered in~\cite{pelletier08} in the context of spectral clustering.  We describe the method of Ng, Jordan and Weiss~\cite{Ng02} for a general kernel $\phi$ in \hbox{Algorithm \ref{algo:NJW}}.  The $K$-means algorithm is initialized with centroids at nearly $90^o$ angles, and then run with only one iteration.  The initial centroids are chosen recursively, starting with any row vector of $\bV$ and then choosing a row vector with largest minimal absolute angles with all the centroids previously chosen.

\begin{algo}[ht]
\caption{Pairwise spectral clustering}
\centering
\begin{tabular}{p{6in}}
\textbf{Input:}\\
\hspace*{.3in} $\{\bx_1,\bx_2,...,\bx_N\} \subset [0,1]^{D}$: the data set\\
\hspace*{.3in} $\eps$: affinity scale\\
\hspace*{.3in} $K$: the number of clusters\\

\textbf{Output:}\\ 
\hspace*{.3in} A partition of the data into $K$ disjoint clusters\\[.1in]

\textbf{Steps:}\\ 
{\bf 1:} Compute the affinity matrix $\bW \in \reals^{N \times N}$, with $W_{ij} = \alpha(\bx_i, \bx_j)$. \\
{\bf 2:} Compute the degree matrix $\mathbf{D}=\diag\{\mathbf{W}\cdot\mathbf{1}\}$, and $\mathbf{Z}=\bD^{-{1}/{2}} \bW \bD^{-{1}/{2}}$. \\
{\bf 3:} Extract $\bU = [\bu_1, \dots, \bu_K]$, orthogonal eigenvectors of $\bZ$ for its $K$ largest eigenvalues. \\
{\bf 4:} Renormalize each {\it row} of $\mathbf{U}$ to have unit norm and let $\bV$ denote the resulting matrix. \\
{\bf 5:} Apply $K$-means to cluster the row vectors of $\mathbf{V}$ in $\mathbb{R}^K$. \\
{\bf 6:} Accordingly group the original points into $K$ disjoint clusters. \\
\end{tabular}
\label{algo:NJW}
\end{algo}

\begin{theorem}
\label{th:NJW}
Consider the generative model described in Section \ref{sec:setting} with surfaces $S_1, \dots, S_K \in \cS(\kappa)$.  Let $\omega_N$ be such that $N^q \phi(\omega_N) = o(1)$ for any $q > 0$.  Assume $\delta - 2 \tau \geq \omega_N \eps$ and that (\ref{eq:N-cond0}) holds.  Then, Algorithm \ref{algo:NJW} is perfectly accurate with high probability.
\end{theorem}
The proof of Theorem \ref{th:NJW} is in Section \ref{proof:NJW}.
We see that Theorem \ref{th:NJW} is very similar to Theorem \ref{th:CC}; for example, with the Gaussian kernel, the separation condition is $\delta - 2 \tau \gg \eps \sqrt{\log N}$.  The respective proofs are essentially parallel as well, though for the latter we follow the outline provided in~\cite{Ng02}.  Thus in theory and under our model, Algorithms \ref{algo:CC} and \ref{algo:NJW} operate under similar conditions.  In practice, however, it is well-known that Algorithm \ref{algo:CC} is substantially more sensitive to the specification of the scale parameter $\eps$.

\subsection{Single linkage clustering}
\label{sec:single}

In the setting of Section \ref{sec:setting}, there is no hope for hierarchical clustering methods using complete or average linkage unless the clusters are separated by a distance comparable to their diameter, or larger.  This is the classical setting, where the goal is typically to form clusters with small diameter~\cite{1090312}.  On the other hand, the ``chaining" property of hierarchical clustering with single linkage is desirable in our context, especially if the cluster is truly lower-dimensional (e.g. generated by sampling near a curve).  In fact, if we stop the procedure whenever the closest distance between clusters exceeds $\eps$, the resulting algorithm is equivalent to Algorithm \ref{algo:CC} with kernel $\phi(s) = {\bf 1}\{s \leq 1\}$.  The procedure is described in Algorithm \ref{algo:single}. 

\begin{algo}[ht]
\caption{Single linkage clustering}
\centering
\begin{tabular}{p{6in}}
\textbf{Input:}\\
\hspace*{.3in} $\{\bx_1,\bx_2,...,\bx_N\} \subset \mathbb{R}^{D}$: the data set\\
\hspace*{.3in} $\eps$: maximum merging distance \\

\textbf{Output:}\\ 
\hspace*{.3in} A partition of the data into disjoint clusters\\[.1in]

\textbf{Steps:}\\
{\bf 0:} Set each point to be a cluster. \\ 
{\bf 1:} Recursively merge the two closest clusters in terms of minimal distance. \\
{\bf 2:} Stop when the distance between any pair of clusters exceeds $\eps$. \\

\end{tabular}
\label{algo:single}
\end{algo}

\begin{corollary}
Under the conditions of Theorem \ref{th:CC}, Algorithm \ref{algo:single} is perfectly accurate with high probability.
\end{corollary}

We mention the paper of Achlioptas and McSherry~\cite{ach05}, which introduces an algorithm based on a combination of spectral clustering and single linkage clustering.  Their analysis shows that their algorithm performs comparatively well in the classical setting.

\section{Optimality in terms of separation between clusters}
\label{sec:sep}

From Theorem \ref{th:CC} we see that Algorithm \ref{algo:CC} is able to correctly identify clusters separated by a distance in the order of the term on the right hand side of (\ref{eq:N-cond0}).  In the classical setting, Algorithm \ref{algo:CC} is accurate when $\delta \geq C \tau$, with $C > 2$; this is valid in any dimension, as explained in Section \ref{sec:D-large}.  The requirement is therefore comparable, actually weaker, than the lower bound achieved in~\cite[Th. 6]{ach05}.  This is assuming we can select an appropriate scale, which we do in Section \ref{sec:scale}.  Note that the algorithm of Achlioptas and McSherry~\cite{ach05} requires selecting the correct number of clusters.

In our framework, the degree of separation required by Algorithm \ref{algo:CC} to be perfectly accurate is close to optimal when the noise level $\tau$ is small, specifically,
$$\tau \prec \min_{k = 1, \dots, K} \diam(S_k) (\log(N_k)/N_k)^{1/d_k}.$$

\begin{theorem}
\label{th:sep}
For any clustering method and any probability $p \in (0,1)$, there are surfaces $S_1, S_2 \in \cS_d(\kappa)$ of diameter at least $1/2$ and separated by $\delta$, with $\delta - 2 \tau \succ (1/N)^{1/d}$, such that, in the context of the generative model of Section \ref{sec:setting}, the method makes at least one mistake with probability at least $p$. 
\end{theorem}
The proof of Theorem \ref{th:sep} is in Section \ref{proof:sep}.

{\it Remark.}
We avoided the case of surfaces of mixed dimensions since the use of more sophisticated tools, such as local density or dimension estimation~\cite{Haro06,levina-bickel}, could possibly narrow the separation.

The conclusion of Theorem \ref{th:sep} is rather weak, though, as it does not give conditions under which any clustering method has a substantial error rate (in terms of labeling the points).  In dimensions one and two, we are able to prove such a result.  In fact, we show that Algorithm \ref{algo:CC} achieves the optimal separation rate, up to a constant factor in dimension one and up to a poly-logarithmic factor in dimension two.  We were not able to prove such a result in higher dimensions.

\begin{theorem}
\label{th:1-sep}
For any clustering method, there are surfaces $S_1, S_2 \in \cS_1(\kappa)$ of diameter at least $1/8$ and separated by $\delta$, with $\delta - 2 \tau \succ \log(N)/N$, on which the method has an error rate exceeding $1/9$ with high probability. 
\end{theorem}
The proof of Theorem \ref{th:1-sep} is in Section \ref{proof:1-sep}.

\begin{theorem}
\label{th:2-sep}
For any clustering method, there are surfaces $S_1, S_2 \in \cS_2(\kappa)$ of diameter at least $1/8$ and separated by $\delta$, with $\delta - 2 \tau \succ 1/(N \log(N) \sqrt{\log \log(N)})$, on which the method has an error rate exceeding $1/9$ with high probability. 
\end{theorem}
The proof of Theorem \ref{th:2-sep} is in Section \ref{proof:2-sep}.

\section{Optimality in terms of robustness}
\label{sec:detection}

\subsection{Dealing with outliers}
\label{sec:outliers}

So far we only considered the case where the data is devoid of outliers.  We now assume that some outliers may be included in the data.  The outliers are sampled from a distribution $\Psi_0$ with density $\psi_0$ with respect to the uniform measure on $[0,1]^D \setminus \bigcup_k B(S_k,\delta)$, again with $\kappa^{-1} \leq \psi_0 \leq \kappa$.  We assume this region is of $D$-volume bounded below by $\kappa^{-1}$.  We denote by $N_0$ the number of outliers.  We highlight the fact that outliers are away from surfaces by at least $\delta$, the same lower bound on the distance that separates two distinct surfaces.

The algorithms considered here are based on pairwise distances, so we need to assume that the outliers are not as densely sampled as the actual clusters, for otherwise they will be indistinguishable from non-outlier points. 

\begin{proposition}
\label{prop:single-outliers}
Assume the conditions of Theorem \ref{th:CC} hold, now in a setting that includes outliers, and, in addition, that $\eps \ll N_0^{-2/D}$.  Then, Algorithm \ref{algo:single} is perfectly accurate with high probability if, when the algorithm stops, singletons are labeled as outliers.
\end{proposition}
The proof of Proposition \ref{prop:single-outliers} is in Section \ref{proof:single-outliers}.

Algorithms \ref{algo:CC} and \ref{algo:NJW} need to be modified in order to deal with outliers.  We introduce an additional step which consists in discarding the data points with low connectivity in the neighborhood graph.  This approach to removing outliers is very natural and was proposed in other works, such as~\cite{spectral_applied,1519716}.  Specifically, fix a sequence $\omega_N \to \infty$ such that $\omega_N \ll \log N$; then, between steps 1 and 2, compute the degree matrix $\bD = \diag(\bW \cdot {\bf 1})$ and discard the points with degree $D_i \leq \omega_N N \eps^D + \log N$.

\begin{lemma} \label{lem:vol1}
For $S \in \cS_d(\kappa)$ and $\eps, \tau > 0$, and $\bx \in B(S,\tau)$,
$$
\vol_D(B(S,\tau) \cap B(\bx, \eps)) \asymp \gamma(S, \tau, \eps),
$$
where
$$
\gamma(S, \tau, \eps) = (\tau \wedge \eps)^{D-d} ((\tau \wedge \eps) \vee (\diam(S) \wedge \eps))^d.
$$
\end{lemma}
The proof of Lemma \ref{lem:vol1} is in Section \ref{proof:vol1}.
Let $\gamma(S, \tau) = \tau^{D-d} (\tau \vee \diam(S))^d$; by Lemma \ref{lem:vol1}, $\vol_D(B(S,\tau) \asymp \gamma(S, \tau)$.

\begin{proposition}
\label{prop:outliers}
Assume the conditions of Theorem \ref{th:CC} hold, now in a setting that includes outliers.  Suppose: 
\begin{equation} \label{eq:N-cond-outliers}
N_k \frac{\gamma(S_k, \tau, \eps)}{\gamma(S_k, \tau)} \gg \omega_N N \eps^D + \log N, \quad \forall k = 1, \dots, K.
\end{equation}
Then, Algorithms \ref{algo:CC} and \ref{algo:NJW} (modified) are both perfectly accurate with high probability.
\end{proposition}
The proof of Proposition \ref{prop:outliers} is in Section \ref{proof:outliers}.
With enough separation as assumed here, outliers are disconnected from non-outliers, and their degree is of order roughly $N \eps^D$.  Therefore, they should be properly identified by the thresholding procedure.  As for non-outliers, the term on the left hand side of (\ref{eq:N-cond-outliers}) is the order of magnitude of the degree of points sampled from $B(S_k, \tau)$, so that (\ref{eq:N-cond-outliers}) essentially guaranties that non-outliers survive the thresholding step.  

In~\cite{1519716} outliers are sampled anywhere in space but away from clusters, which corresponds to $\Psi_0$ having support $[0,1]^D \setminus \bigcup_k B(S_k, \tau)$.  In that case, perfect accuracy is impossible, as the algorithms will confuse outliers within $\eps$ from $B(S_k, \tau)$ with points belonging to $\cX_k$.  However, knowing that, with high probability, there are at most $O(N_0 \eps)$ such outliers (in fact $O(N_0 \eps^{D-d_k})$ if $\tau \prec \eps$, by Lemma \ref{lem:vol1}), the algorithms make a mistake on a negligible fraction of outliers.

\subsection{Clustering at the detection threshold}

Assume each cluster is sufficiently sampled, which we rigorously define as:
\begin{equation} \label{eq:N-cond-lb}
N_k \geq (N^{d_k/D} \vee N \tau^{D-d_k}) \log(N), \quad \forall k = 1, \dots, K.
\end{equation}
Note that the related condition $N_k \asymp N^{d_k/D}$ is equivalent to requiring that, within each cluster, the distance between a point and its nearest-neighbor is of order $N^{-1/D}$.  With (\ref{eq:N-cond-lb}) holding, the choice $\eps = (\log N/N)^{1/D}$ implies both (\ref{eq:N-cond0}) and (\ref{eq:N-cond-outliers}), so that Algorithms \ref{algo:CC} and \ref{algo:NJW} (modified) are perfectly accurate with high probability, even in a setting including outliers.

Now, instead of clustering, consider the task of detecting the presence of a cluster hidden among a large number of outliers.  We observe the data, $\bx_1, \dots, \bx_N$, and want to decide between the following two hypotheses: under the null, the points are all outliers; under the alternative, there is a surface $S \in \cS_d$ such that $N_1$ points are sampled from $B(S,\tau)$, while the rest of the points, $N-N_1$ of them, are sampled as outliers.
Assuming that the parameters $d$ and $\tau$ are known, it is shown in~\cite{CTD,ery-thesis} that the scan statistic is able to separate the null from the alternative if 
$$N_1 \gg N^{d/D} \vee N \tau^{D-d}.$$ 
The author is not aware of a method that improves on those rates, and from translating recent results on detection in graphs~\cite{MR2435454}, there is evidence that those rates are optimal up to a poly-logarithmic factor.  This condition is essentially the same as (\ref{eq:N-cond-lb}), except for the $\log N$ factor.  Hence, Algorithms \ref{algo:CC} and \ref{algo:NJW} (modified) solve the clustering task perfectly within a poly-logarithmic factor of the best known signal-to-noise ratio required for the detection task.


\section{Discussion}
\label{sec:discussion}

\subsection{Selecting the scale parameter}
\label{sec:scale}

Choosing the affinity scale $\eps$ is critical in all algorithms described here, and more generally in any method which uses a neighborhood graph.  Assuming (\ref{eq:N-cond-lb}) holds, we already saw that the choice $\eps = (\log N/N)^{1/D}$ implies (\ref{eq:N-cond0}), so that, with enough separation, the Algorithms \ref{algo:CC}, \ref{algo:NJW} and \ref{algo:single} are accurate.  In terms of separation, this allows the clusters to be as close as $(\log N/N)^{1/D}$.  As seen in Section \ref{sec:sep}, this is not optimal.  Though the choice of $\eps$ may be made more precise with more information on the clusters, like the number of points sampled from them and their dimension, this information may not be available.

In practice, choosing the scale is still an ongoing line of research, with similarities with bandwidth selection in kernel smoothing.  We focus on the local scaling method of Zelnik-Manor and Perona~\cite{Zelnik-Manor04}, where $\alpha(\bx_i, \bx_j)$ is defined as $\phi(\|\bx_i - \bx_j\|/\sqrt{\eps_i  \eps_j})$, with $\eps_i$ equal to the distance between $\bx_i$ and its $\ell$th nearest neighbor.  When $\phi$ is of compact support, this essentially means that if $\bx_i$ is not among the first $\ell$ nearest neighbors of $\bx_j$ and {\it vice versa}, $\bx_i$ and $\bx_j$ are not connected in the neighborhood graph, corresponding to a mutual $k$-nearest neighbor graph.  The parameter $\ell$ replaces $\eps$ as the tuning parameter, effectively setting the number of neighbors (degree) instead of the neighborhood range.  This allows the scaling to adapt to the local sampling density.  

\begin{proposition}
\label{prop:NN}
Consider the generative model of Section \ref{sec:setting} with surfaces $S_1, \dots, S_K \in \cS(\kappa)$.  In terms of separation, for a sequence $\omega_N \to \infty$ such that $\omega_N = o(\log N)$, assume that 
$$
\delta - 2 \tau \gg \max_{k = 1, \dots, K} \max \left\{\begin{array}{l}
(\tau \vee \diam(S_k)) (\omega_N \log(N)/N_k)^{1/d_k}, \\   
(\tau \vee \diam(S_k))^{d_k/D} \tau^{1 - d_k/D} (\omega_N \log(N)/N_k)^{1/D}
\end{array}\right\}.
$$
Then, the local scaling version of Algorithm \ref{algo:CC} with $\ell = \omega_N \log N$ is perfectly accurate with high probability.
\end{proposition}
The proof of Proposition \ref{prop:NN} is in Section \ref{proof:NN}.
As a consequence of Proposition \ref{prop:NN}, with local scaling, Algorithm \ref{algo:CC} essentially achieves the separation in (\ref{eq:N-cond0}).  So in that sense local scaling offers a (near-)optimal way of building the neighborhood graph.

A weaker result, directly dealing with a $k$-nearest neighbor graph and without the optimality implications on the amount of separation, appears in Brito, Ch‡vez, Quiroz and Yukich~\cite{RePEc:eee:stapro:v:35:y:1997:i:1:p:33-42}.  They find that, when the separation between clusters remains fixed, choosing $k$ of order $\log N$ makes Algorithm \ref{algo:CC} work.  
However, assuming the underlying surfaces have diameter of order 1 and same dimension $d$, Maier, Hein and von Luxburg~\cite{1519716} find that the optimal $k$ is of order $N \delta^d$, which is of order $\log N$ only when $\delta$ is of order $(\log(N)/N)^{1/d}$.  As they point out in their paper, it makes sense to use a larger $k$ if the separation between clusters is large.  However, it is still not clear how to automatically choose an optimal $k$ without information on the separation between clusters.

\subsection{Selecting the number of clusters}
\label{sec:K}

Algorithm \ref{algo:NJW} depends on choosing the number of clusters $K$ appropriately.  Since the method relies on the few top eigenvectors of the matrix $\bZ$, a first approach consists in choosing $K$ by inspecting the eigenvalues of $\bZ$.  We provide below an estimate for the gap between the $\lambda_K(\bZ)$ and $\lambda_{K+1}(\bZ)$, which in theory may be used to select the correct number of clusters.  Note that the bound we derive is very crude; for example, if the surfaces are affine subspaces and the sampling is exact ($\tau = 0$), a sharper bound of order $\eps^2$ holds~\cite{boyd}.

\begin{proposition}
\label{prop:K-choice}
Under the conditions of Theorem \ref{th:NJW}, $\lambda_K(\bZ) - \lambda_{K+1}(\bZ) \succ N^{-2}$  with high probability, 
\end{proposition}
The proof of Proposition \ref{prop:K-choice} is in Section \ref{proof:K-choice}.
In practice, this method is seen to work poorly; for example, in~\cite{4517734}, choosing the number of clusters by cross-validation is observed to be more reliable.  In~\cite{Zelnik-Manor04}, the authors suggest examining the few top eigenvectors instead of the eigenvalues.  We do not study these methods here.  In a slightly different context, Biau, Cadre and Pelletier~\cite{pelletier07} propose essentially to count the number of connected components found by Algorithm \ref{algo:CC}.  The conditions stated in Theorem \ref{th:CC} of course guarantee this estimate is accurate with high probability.  Their result is however more precise.

\subsection{When the ambient dimension is large}
\label{sec:D-large}

In a number of modern applications, such as clustering of gene expression data~\cite{gene-mst,299448}, document retrieval~\cite{312186,775110} or clustering 3D objects in computer vision~\cite{Ho03}, the ambient dimension $D$ is routinely several orders of magnitude larger than the number of points $N$.
Though we can always restrict ourselves to the subspace where the points live, which is of dimension $N-1$ or less, we consider here the situation where the ambient space is the unit ball in an infinite-dimensional space, for example a Hilbert space as considered in~\cite{4439834}.  As defining a uniform distribution in such a space is a non-trivial endeavor~\cite{MR1479180}, we modify the model slightly.  We assume that the points are generated from the surface $S$ as follows: $\bx_i = \by_i + \tau \bz_i$, where $\by_i \sim \Psi_S$, a probability measure equivalent to the uniform measure on $S$, and $\bz_i \sim \Psi_B$, a probability measure with support in the unit ball.  Outliers are directly sampled from $\Psi_B$.

Under this setting, Theorems \ref{th:CC} and \ref{th:NJW} remain valid in the case where $\tau \prec \eps$, where the condition (\ref{eq:N-cond0}) does not involve the ambient dimension:
$$
\eps \gg \max_{k = 1, \dots, K} \diam(S_k) (\log(N_k)/N_k)^{1/d_k}.
$$
The arguments are essentially identical.  The case $\eps \prec \tau$ is not as straightforward, since this is the regime where, in some sense, the effective dimension of $B(S_k, \tau)$ is the ambient dimension, and the specifics of the distribution $\Psi_B$ come into play.  Also, our arguments involve using packings of $B(S_k, \tau)$, so that the actual structure of the ambient space is critical.  The same comments apply for the case where outliers are present in the data.

\subsection{Computational Issues}
\label{sec:computations}

We consider the computational complexity of each of the methods described earlier in the paper.  Below, $\beta$ is a large enough constant.  

Building the neighborhood graph may be done by brute force in $O(\rho N^2)$ flops, where $\rho$ is the cost of computing the distance between two points; for example, without further structure, $\rho \asymp D$ in dimension $D$.  This may be done more effectively using an algorithm for range search, or $\ell$-nearest neighbor search for the local scaling version.  In low dimensions, $D \prec \log \log N$, this may be done with kd-trees in $O(N \log(N)^\beta)$ flops.  In higher dimensions, other alternatives may work better~\cite{1290769}. 

Once the neighborhood graph is built, Algorithm \ref{algo:CC} extracts the connected components of the graph, which may be done in $O(\omega_N N \log N)$ flops if using the local scaling version with $\ell = \omega_N \log N$ as suggested in Section \ref{sec:scale}, since in that case the maximum degree is not larger than $\ell$.  Algorithm \ref{algo:NJW} extracts the leading eigenvectors of $\bZ$, which may be done in $O(K N \log(N)^\beta)$ flops, using Lanczos-type algorithms~\cite{MR1948689} since, using again local scaling with $\ell = \omega_N \log N$, $\bZ$ has about $\omega_N \log N$ non-zero coefficients per row.  So in both Algorithms \ref{algo:CC} and \ref{algo:NJW} with local scaling, the total computational complexity is $O(N \log(N)^\beta)$ flops in low dimensions; and at most $O(\rho N^2 + N \log(N)^\beta)$ flops in higher dimensions.  Algorithm \ref{algo:single} runs in $O(\rho N^2 \log(N)^\beta)$ flops in any dimension~\cite{And02hierarchicalclustering}.



\section{Proofs}
\label{sec:proofs}

In all the proofs that follow, we assume for concreteness that the different sampling distributions  are in fact uniform distributions over their respective support and that the kernel $\phi$ has support $[0,1]$; we assume that $\tau > 0$ and that all underlying surfaces are of diameter of order 1.  The remaining cases may be treated similarly.  We use $C$ to denote a generic positive constant, whose actual value may change from place to place.

\subsection{Proof of Theorem \ref{th:CC}}
\label{proof:CC}

First, two distinct clusters $\cX_k$ and $\cX_\ell$ are disjoint in the graph.  Indeed, for $i \in I_k$ and $j \in I_\ell$, $\|\bx_i - \bx_j\| \geq \delta - 2 \tau > \eps$ and therefore $\alpha(\bx_i, \bx_j) = 0$ since $\phi$ is supported in $[0,1]$.  

Now consider a single cluster of size $N$ generated from a surface $S \in \cS_d$.
Let $\by_1, \dots, \by_n$ be an $\eps/5$-packing of $B(S,\tau)$.  Because $B(S,\tau) \subset \bigcup_j B(S,\tau) \cap B(\by_j, \eps/5)$, $\vol_D(B(S,\tau)) \leq \sum_j \vol_D(B(S,\tau) \cap B(\by_j, \eps/5))$, so that by Lemma \ref{lem:vol1}, $\tau^{D-d} \prec n \eps^d (\eps \wedge \tau)^{D-d}$; on the other hand, $\bigsqcup_j B(S,\tau) \cap B(\by_j, \eps/10) \subset B(S,\tau)$ implies $\sum_j \vol_D(B(S,\tau) \cap B(\by_j, \eps/10)) \leq \vol_D(B(S,\tau))$, so that by Lemma \ref{lem:vol1} again, $n \eps^d (\eps \wedge \tau)^{D-d} \prec \tau^{D-d}$.  Hence, $n \asymp (\eps \vee \tau)^{D-d} \eps^{-D}$.

By condition (\ref{eq:N-cond0}), $N \gg n \log n$, and so by a simple modification of Lemma 2 in \cite{balls-in-bins}, with high probability each ball $B(\by_j, \eps/5)$ contains at least one (in fact, order $N/n$) data point(s).  If $\cX$ were disconnected, we could group the data points into two groups in such a way that the minimum distance between the two groups would exceed $\eps$.  By the triangle inequality, this would imply a grouping of the $\by_j$'s into two groups with a minimum pairwise distance of $(3/5)\eps$.  The balls $B(\by_j, \eps/5), j=1,\dots,n$, would then be divided into two disjoint groups, which contradicts the fact that they cover the connected set $S$.

\subsection{Proof of Theorem \ref{th:NJW}}
\label{proof:NJW}

We follow the strategy outlined in~\cite{Ng02} based on verifying the following conditions (where (A4) has been simplified).
For $k = 1, \dots, K$, let $\bW_k$ denote the submatrix of $\bW$ corresponding to the index set $I_k$.  For $i \in I_k$, define $\rD_i = \sum_{\bx_j \in \cX_k} W_{ij}$, which is the degree of $\bx_i$ within the cluster it belongs to.  Let $\bv_1, \dots, \bv_N$ denote the row vectors of $\bV$.

\begin{itemize}
\item[] {\bf (A1)}  For all $k = 1, \dots, K$, the second largest eigenvalue of $\bW_k$ is bounded above by $1 - \zeta$.
\item[] {\bf (A2)}  For all $k, \ell = 1, \dots, K$, with $k \neq \ell$, 
$$\sum_{i \in I_k} \sum_{j \in I_\ell} \frac{W_{ij}^2}{\rD_{i} \rD_{j}} \leq \nu_1.$$
\item[] {\bf (A3)}  For all $k = 1, \dots, K$ and all $i \in I_k$,
$$\frac{1}{\rD_i} \sum_{j \notin I_k} W_{ij} \leq \nu_2 \left(\sum_{s, t \in I_k} \frac{W_{st}^2}{\rD_{s} \rD_{t}}\right)^{-1/2}.$$
\item[] {\bf (A4)}  For all $k = 1, \dots, K$ and all $i, j \in I_k$, $\rD_i \leq \theta \rD_j$.
\end{itemize}

We present below a slightly modified version of Theorem 2 in~\cite{Ng02}.  

\begin{theorem}[Based on Th. 2 in~\cite{Ng02}]
\label{th:NJW-orig}
Under (A1)-(A4), there is an orthonormal set $\{\br_1, \dots, \br_K\} \subset \bbR^K$ such that,
$$\sum_{k = 1}^K \sum_{i \in I_k} \|\bv_i - \br_k\|^2 \leq 8 \theta \zeta^{-2} (K^2 \nu_1 + K \nu_2^2) N.$$ 
\end{theorem}
The proof of Theorem \ref{th:NJW-orig} is in Section \ref{proof:NJW-orig}.  It is partly based on information that Andrew Ng shared with the author and the proof of~\cite[Th 4.5]{spectral_theory} by Chen and Lerman.  Note that the latter deals with the special case where the clusters are of comparable sizes ($N_k \asymp N$) and of same dimension ($d_k = d$), and the result they obtain is somewhat different.

We show below that $\zeta \succ N^{-2}$, that $\nu_1, \nu_2 \prec N^{-q}$ for any $q > 0$, and that $\theta \prec 1$.  Hence, the right hand side in the expression above is of order $N^{-q}$ for any $q > 0$, since $K$ is assumed fixed.  Hence, $\max_i \min_k \|\bv_i - \br_k\| \to 0$ and therefore, since the $\br_k$'s are themselves orthonormal, $K$-means with mear-orthogonal initialization outputs the perfect clustering with high probability.  We now turn to verifying (A1)-(A4), in reverse order.

{\bf (A4):} We show that, with high probability and uniformly over $i, j \in I_k$,  
$$\rD_i \asymp \rD_j \asymp N_k \eps^{D} / (\eps \vee \tau)^{D-d_k}.$$
Consider a single cluster of size $N$, generated from sampling near a surface $S$ of dimension $d$.  Assume that (\ref{eq:N-cond0}) holds, namely $N \eps^{D} / (\eps \vee \tau)^{D-d} \gg \log N.$  Given $\bx_i$, $\alpha(\bx_i, \bx_j), j \neq i$ are i.i.d. random variables in $[0,1]$, with mean $\xi_i \asymp \Psi(B(\bx_i, \eps))$.  Note that $\xi_i \asymp \eps^{D} / (\eps \vee \tau)^{D-d}$ by Lemma \ref{lem:vol1}.  Using Hoeffding's Inequality, in the form of inequality (2.1) in~\cite{hoeffding}, there is a constant $C > 0$, such that
$$\pr{|D_i - N \xi_i| > (1/2) N \xi_i} \leq 2 \exp(- C \ N \xi_i).$$
By (\ref{eq:N-cond0}), $N \xi_i  \gg \log N$ so that, with Boole's Inequality, we conclude that $D_i \asymp N \xi_i$ uniformly over all $i$, with high probability.

{\bf (A3):} Fix $k = 1, \dots, K$; by applying the result in (A4) with $\phi^2$ as a kernel, we get the following order of magnitude, uniformly over $s \in I_k$, 
$$\sum_{t \in I_k} W_{st}^2 \asymp N_k \eps^{D} / (\eps \vee \tau)^{D-d_k}.$$
Therefore, by (A4) and then (\ref{eq:N-cond0}),
$$\sum_{s, t \in I_k} \frac{W_{st}^2}{\rD_{s} \rD_{t}} \asymp (\eps \vee \tau)^{D-d_k} \eps^{-D} \ll N.$$
Now, take two points, $\bx_i \in \cX_k$ and $\bx_j \notin \cX_k$.  Because $\|\bx_i - \bx_j\| \geq \delta - 2 \tau \geq \omega_N \eps$, we get $W_{ij} \leq \phi(\omega_N)$.  With (A4) and (\ref{eq:N-cond0}), this implies that
$$\frac{1}{\rD_i} \sum_{j \notin I_k} W_{ij} \leq (\eps \vee \tau)^{D-d_k} \eps^{-D} \phi(\omega_N) \ll N \phi(\omega_N).$$
Therefore, we can take $\nu_2 = N^{3/2} \phi(\omega_N)$, so that $\nu_2 \prec N^{-q}$ for any $q > 0$.
  
{\bf (A2):} We apply the same arguments we just used to bound the sum on the left hand side of (A3).  In particular, we can take $\nu_1 = N^{2} \phi(\omega_N)^2$, so that $\nu_1 \prec N^{-q}$ for any $q > 0$.

{\bf (A1):} We prove that the spectral gap $\zeta$ satisfies $\zeta \succ N^{-2}$ with high probability.  As suggested in \cite{Ng02}, we approach this through a lower bound on the Cheeger constant.  Consider a single cluster of size $N$, generated from sampling near a surface of dimension $d$.  Assume that (\ref{eq:N-cond0}) holds, namely $N \eps^{D} / (\eps \vee \tau)^{D-d} \gg \log N.$  That $\bZ$ has eigenvalue 1 with multiplicity 1 results from the graph being fully connected.  The Cheeger constant of $\bW$ is defined as:
$$h = \min_{|I| \leq N/2} \frac{\sum_{i \in I} \sum_{j \notin I} W_{ij}}{\sum_{i \in I} D_i},$$
where the minimum is over all subsets $I \subset \{1, \dots, N\}$ of size $|I| \leq N/2$.
The spectral gap of $\bZ$ is then of order at least $h^2$.  Using (A4), we get the lower bound:
$$h \succ (N \eps^{D} / (\eps \vee \tau)^{D-d})^{-1} \ \min_{|I| \leq N/2}  \frac{\sum_{i \in I} \sum_{j \notin I} {\bf 1}\{\| \bx_i - \bx_j \| \leq \eps\}}{|I|}.$$
Let $\by_1, \dots, \by_n$ be an $\eps/5$-packing of $B(S,\tau)$ and define let $A_j = B(S,\tau) \cap B(\by_j, \eps/5)$.  Not only are the cells $A_1, \dots, A_n$ non-empty with high probability, actually they all contain order $N \eps^{D} / (\eps \vee \tau)^{D-d}$ points; see Lemma 2 in \cite{balls-in-bins} or the proof of (A4).  For a fixed $I$ such that $|I| \leq N/2$, there are necessarily two cells $A_j$ and $A_\ell$ with $\|\by_j - \by_\ell\| \leq \eps/2$ such that $\# (A_j \cap \{\bx_i: i \in I\}) \leq \# A_j/2$ and  $\# (A_\ell \cap \{\bx_i: i \in I\}) \geq \# A_\ell/2$, for otherwise it would imply that $S$ is disconnected.  Therefore, given that points in $A_j$ and $A_\ell$ are within distance $\eps$, and that both $A_j$ and $A_\ell$ contain order $N \eps^{D} / (\eps \vee \tau)^{D-d}$ points, we have  
$$\sum_{i \in I} \sum_{j \notin I} {\bf 1}\{\| \bx_i - \bx_j \| \leq \eps\} \succ N \eps^{D} / (\eps \vee \tau)^{D-d}.$$
As this is true for any $I$ such that $|I| \leq N/2$, we have that $h \succ N^{-1}$.

\subsection{Proof of Theorem \ref{th:sep}}
\label{proof:sep}

We start with the one dimensional case, which is substantially simpler than the situation in higher dimensions, as the boundaries of one-dimensional sets are just points.  
We work with the supnorm for convenience and clarity.  For two probability distributions $P, Q$, let $H(P,Q)$ denote their Hellinger distance~\cite[Chap. 13]{MR2135927}.  

\subsubsection{The case $d = 1$}

Consider the line segment within $[0,1]^D$ generated by the first canonical vector, which we identify with $[0,1]$.  For $f \in (0,1)$ and $\delta > 0$, define
$$S_1 = \{u \in [0,1]: u < f - \delta/2\};$$
$$S_2 = \{u \in [0,1]: u > f + \delta/2\}.$$
For $\tau \in (0, \delta/2)$, generate a cluster $\cX_1$ (resp. $\cX_2$) by sampling uniformly from $B(S_1, \tau)$ (resp. $B(S_2, \tau)$), where
$$B(S_1, \tau) = \{(u, \ba) \in [0,1] \times [0,\tau]^{D-1}: u < f - \delta/2 + \tau\};$$
$$B(S_2, \tau) = \{(u, \ba) \in [0,1] \times [0,\tau]^{D-1}: u > f + \delta/2 + \tau\}.$$
The sampling is in proportion with the volume of these regions, i.e. $N_j \propto \vol_D(B(S_j, \tau))$.
By sufficiency, we need only consider the first coordinate, effectively reducing the case to that of $\tau = 0$.  From this perspective, the setting is that of points sampled from $P_{f, \delta}$, the uniform distribution on $[0,1] \setminus [f - \delta/2, f + \delta/2]$.

Let $f_0 = 1/4$ and $f_1 = 3/4$, and assume $\delta < 1/4$.  Suppose we want to decide between $P_{f_0, \delta}^{\otimes N}$ and $P_{f_1, \delta}^{\otimes N}$.  From a clustering method, we obtain a test in the following way: after grouping the points, we reject the null hypothesis if $f_1$ separates the two clusters.  Since the interval $[3/8, 5/8]$ contains more than $N/5$ data points with high probability, the clustering method has an error rate of at least $1/5$ when as a test it makes an error.  Fix a probability $p \in (0,1)$.  As  a consequence of~\cite[Th. 13.1.3]{MR2135927}, and 
$$
N H^2(P_{f_0, \delta}, P_{f_1, \delta}) = N \delta/(2 - 2 \delta) \prec N \delta,
$$
any test makes an error with probability at least $p$ if $N \delta$ is small enough.

\subsubsection{The case $d \geq 2$}

Consider the $d$-dimensional affine surface within $[0,1]^D$ generated by the first $d$ canonical vectors, which we identify with $[0,1]^d$.  For a function $f: [0,1]^{d-1} \to [0,1]$ and $\delta > 0$, define
$$S_1 = \{(\bu, v) \in [0,1]^{d-1} \times [0,1]: v \leq f(\bu) - \delta/2\};$$
$$S_2 = \{(\bu, v) \in [0,1]^{d-1} \times [0,1]: v \geq f(\bu) + \delta/2\}.$$
For $\tau \in (0, \delta/2)$, generate a cluster $\cX_1$ (resp. $\cX_2$) by sampling uniformly from $B(S_1, \tau)$ (resp. $B(S_2, \tau)$), where
$$
B(S_1, \tau) = \{(\bu, v, \ba) \in [0,1]^{d-1} \times [0,1] \times [0,\tau]^{D-d}: v \leq f(\bu) - \delta/2 + \tau\};
$$
$$
B(S_2, \tau) = \{(\bu, v, \ba) \in [0,1]^{d-1} \times [0,1] \times [0,\tau]^{D-d}: v \geq f(\bu) + \delta/2 + \tau\}.
$$
Again, the sampling is in proportion with the volume of these regions.
By sufficiency, we need only consider the first $d$ coordinates, effectively reducing the case to that of $\tau = 0$.  Henceforth, the setting is that of points sampled from $P_{f, \delta}$, the uniform distribution on $[0,1]^d \setminus R_{f, \delta}$, where 
$$R_{f, \delta} = \{(\bu, v) \in [0,1]^{d-1} \times [0,1]: f(\bu) - \delta/2 < v < f(\bu) + \delta/2\}.$$
For $A > 0$, consider the function $f_0(\bu) = 1/2 + \delta$ if $\bu \in [0,A \delta]^{d-1}$ and $f_0(\bu) = 1/2$ otherwise.  For $A$ large enough relative to $\kappa$, we have $S_1, S_2 \in \cS(\kappa)$.  Let $f_1 = 1 - f_0$ and consider testing $P_{f_0, \delta}^{\otimes N}$ versus $P_{f_1, \delta}^{\otimes N}$.
From a clustering method, we obtain a test in the same way: after grouping the points, we reject the null hypothesis if the graph of $f_1$ separates the two clusters.  The region $[0, A \delta]^{d-1} \times [1/2-\delta/2, 1/2 + \delta/2]$ is non-empty with non-negligible probability if $N \delta^d$ is bounded away from zero.  When this happens, the clustering method `misclassifies' the points falling in that region when as a test it makes an error.  Fix a probability $p \in (0,1)$.  As  a consequence of~\cite[Th. 13.1.3]{MR2135927}, and 
$$
N H^2(P_{f_0, \delta}, P_{f_1, \delta}) = A \cdot N \delta^d/(2 - 2 \delta) \prec N \delta^d,
$$
any test makes an error with probability at least $p$ if $N \delta^d$ is small enough.

\subsection{Proof of Theorem \ref{th:1-sep}}
\label{proof:1-sep}

In dimension one, we show that the logarithmic factor is needed.  This seems quite intuitive, since the longest distance between any pair of consecutive points is of order $\log(N)/N$~\cite{minimum-distance}.  We build on the proof of Theorem \ref{th:sep}.
Define $m = \delta^{-1}$, assumed to be an integer for simplicity.  Consider $f_j = j \delta$ for $1 \leq j \leq m$, and define $J_0 = \{j: (1/8) m \leq j \leq (3/8) m\}$ and $J_1 = \{j: (5/8) m \leq j \leq (7/8) m\}$.   Suppose we want to decide between $P_{f_j, \delta}^{\otimes N}, \ j \in J_0$, and $P_{f_j, \delta}^{\otimes N}, \ j \in J_1$.  With $\delta < 1/8$, the clustering method has an error rate of at least $1/9$ when as a test it makes an error.  


\begin{lemma}
\label{lem:1-detect}
Consider testing $P_{f_j, \delta}^{\otimes N}, \ j \in J_0$ versus $P_{f_j, \delta}^{\otimes N}, \ j \in J_1$.  If $\delta \leq C \log(N)/N$ with $C < 1$, then for any test, the sum of the probabilities of type I and type II errors tends to 1.
\end{lemma}
The proof of Lemma \ref{lem:1-detect} is in Section \ref{proof:1-detect}.

\subsection{Proof of Theorem \ref{th:2-sep}}
\label{proof:2-sep}

We build on the proof of Theorem \ref{th:sep}.
Define $m = (A \delta)^{-1}$, assumed to be an integer for simplicity.  For a sequence $\pi_j \in \{-1, 1\}, \ j = 1, \dots, m$, consider the function $f_{0,\pi}(u) = 1/4 + (\pi_1 + \cdots + \pi_j) \delta$ if $u \in [(j-1) A \delta, j A \delta)$, for $1 \leq j \leq m$.  Similarly, define $f_{1, \pi}$ by replacing $1/4$ with $3/4$.  Suppose we want to decide between $P_{f_{0, \pi}, \delta}^{\otimes N}, \pi \in \{-1, 1\}^m$ and $P_{f_{1, \pi}, \delta}^{\otimes N}, \pi \in \{-1, 1\}^m$.  With $A$ large enough and $\delta$ small, we have $1/A + \delta/2 < 1/8$ and the clustering method has an error rate of at least $1/9$ when as a test it makes an error.

\begin{lemma}
\label{lem:2-detect}
Consider testing $P_{f_{0, \pi}, \delta}^{\otimes N}, \pi \in \{-1, 1\}^m$ versus $P_{f_{1, \pi}, \delta}^{\otimes N}, \pi \in \{-1, 1\}^m$.  If 
$$\delta \ll N^{-1/2} \log(N)^{-1} \log\log(N)^{-1/2},$$ 
then for any test, the sum of the probabilities of type I and type II errors tends to 1.
\end{lemma}
The proof of Lemma \ref{lem:2-detect} is in Section \ref{proof:2-detect}.

\subsection{Proof of Proposition \ref{prop:single-outliers}}
\label{proof:single-outliers}

It is enough to show that, with high probability, no two outliers are within distance $\eps$.  So consider $\bx_1, \dots, \bx_{N_0}$ outliers sampled according to $\Psi_0$.  Let $D_i$ be the number of data points in $B(\bx_i, \eps)$ other than $\bx_i$ itself.  Then, $\pr{D_i > 0} = 1 - (1 - \Psi_0(B(\bx_i, \eps)))^{N_0 - 1} \leq 1 - \exp(- C N_0 \eps^D)$, for some constant $C > 0$,  since $\Psi_0(B(\bx_i, \eps)) \asymp \eps^D$ uniformly in $i$ and $\eps$ is small.
Hence, using Boole's inequality, we have
$$
\pr{\exists i: D_i > 0} \leq N_0 (1 - \exp(- C N_0 \eps^D)) \leq 2 C \eps^D N_0^2 = o(1).
$$

\subsection{Proof of Proposition \ref{prop:outliers}}
\label{proof:outliers}

We need to prove that with high probability, all outliers have degree bounded above by $\omega_N N \eps^D  + \log N$ and that all non-outliers have degree exceeding that threshold.  In both cases, the arguments are similar to those used in obtaining (A4) in the proof of Theorem \ref{th:NJW}.

For the first part, consider the case of $N$ outliers.  Given $\bx_i$, $\alpha(\bx_i, \bx_j), j \neq i$ are i.i.d. random variables in $[0,1]$, with mean $\xi_i \asymp \Psi_0(B(\bx_i, \eps))$.  By Hoeffding's Inequality, in the form of inequality (2.1) in~\cite{hoeffding}, we have
$$\pr{D_i > \omega_N N \xi_i + \log N} \leq \exp( - \log(\omega_N) \log(N) ).$$
We conclude using this, the fact that $\xi_i \asymp \eps^{D}$, and Boole's Inequality.  

For the latter, consider a single cluster of size $N$, generated from sampling near a surface $S$ of dimension $d$.  Assume that (\ref{eq:N-cond-outliers}) holds, namely $N \eps^{D} / (\eps \vee \tau)^{D-d} \gg \omega_N N \eps^D + \log N$.  In proving (A4) in the proof of Theorem \ref{th:NJW}, we found that, with high probability, $D_i \asymp N \xi_i \asymp N \eps^{D} / (\eps \vee \tau)^{D-d}$ uniformly over all $i$.  Therefore, with high probability, $D_i > \omega_N N \xi_i + \log N$ for all $i$.

\subsection{Proof of Proposition \ref{prop:NN}}
\label{proof:NN}

It is enough to show that, for each $k = 1, \dots, K$, the distance from $\bx_i$ to its $\ell$th nearest neighbor within $B(S_k, \tau)$, denoted $\eps_i^*$, satisfies 
$$
\eps_i^* \asymp (\omega_N \log(N)/N_k)^{1/d_k} \vee \tau^{1 - d_k/D} (\omega_N \log(N)/N_k)^{1/D}.
$$
Indeed, under the assumed separation, this will show that $\eps_i = \eps_i^*$, which in turn implies that the clusters are disjoint in the neighborhood graph; and also, that each cluster is connected in the neighborhood graph, since (\ref{eq:N-cond0}) is satisfied (at $k$) with $\eps_i^*$ in place of $\eps$ and therefore the arguments for Theorem \ref{th:CC} apply directly.

By (A4) in the proof of Theorem \ref{th:NJW}, we see that, uniformly over $\bx_i \in B(S_k, \tau)$, the number of points sampled from $B(S_k, \tau) \cap B(\bx_i, \eps)$ is of order of magnitude $N_k \eps^D/(\tau \vee \eps)^{D-d_k}$ for $\eps$ satisfying (\ref{eq:N-cond0}), that is
$$
\eps \gg (\log(N_k)/N_k)^{1/d_k} \vee \tau^{1 - d_k/D} (\log(N_k)/N_k)^{1/D}.
$$
In setting $N_k (\eps_i^*)^D/(\tau \vee \eps_i^*)^{D-d_k} \asymp \ell$, with $\ell = \omega_N \log N$, we find $\eps_i^*$ of the right order of magnitude.

\subsection{Proof of Proposition \ref{prop:K-choice}}
\label{proof:K-choice}

We use the notation introduced in the proof of Theorems \ref{th:NJW} and \ref{th:NJW-orig}.  
By Proposition \ref{prop:inf}, $\brZ$ has eigenvalue 1 with multiplicity $K$ and an eigengap bounded below by $\zeta$.  We also know that $\|\bZ - \brZ\|_F \prec \sqrt{\nu_1 + \nu_2^2}$ by Proposition \ref{prop:perturb} (holding $K$ fixed), assuming (A2)-(A3) are satisfied.  We then compare the spectrum of $\bZ$ and $\brZ$ using \cite[Th. IV.4.8]{MR1061154}:
$$\sum_{m=1}^N |\lambda_m(\bZ) - \lambda_m(\brZ)|^2 \leq \|\bZ - \brZ\|_F^2.$$
This implies that
$$\lambda_K(\bZ) - \lambda_{K+1}(\bZ) \geq \lambda_K(\brZ) - \lambda_{K+1}(\brZ) - 2 \|\bZ - \brZ\|_F \geq \zeta - 2 \sqrt{\nu_1 + \nu_2^2}.$$
We then conclude using the order of magnitudes, $\zeta \succ N^{-2}$ and $\nu_1, \nu_2 \prec N^{-q}$, for any $q > 0$, holding with high probability, obtained when verifying (A1)-(A3).

\appendix

\section{Proof of Theorem \ref{th:NJW-orig}}
\label{proof:NJW-orig}

The strategy outlined in the paper of Ng, Jordan and Weiss~\cite{Ng02} consists in first analyzing the case of infinite separation ($\delta = \infty$) and then in treating the case of finite separation as a perturbation of the case of infinite separation, and (A1)-(A4) are used to control the amount of perturbation.  We emphasize that Andrew Ng shared with the author the arguments underlying the proof of Theorem \ref{th:NJW-orig}.  Inspired by the proof of Chen and Lerman~\cite[Th. 4.5]{spectral_theory}, we present below a slightly different proof, made more concise thanks to a recent result by Zwald and Blanchard~\cite[Th. 3]{Zwald06}.

\subsection{The case of infinite separation}

Let $\brW$ denote the similarity matrix in this situation, so that $\rW_{ij} = W_{ij}$ if there is $k$ such that $i, j \in I_k$, and $\rW_{ij} = 0$ otherwise.  Let $\brD, \brZ, \brU, \brV$ denote the matrices obtained from $\brW$ following Algorithm \ref{algo:NJW}.  Let $\bru_1, \dots, \bru_N$, and $\brv_1, \dots, \brv_N$, denote the row vectors of $\brU$ and $\brV$ respectively.

\begin{proposition}
\label{prop:inf}
Assuming (A1) holds, the matrix $\brZ$ has top eigenvalue 1 with multiplicity $K$ and eigengap bounded below by $\zeta$.  Moreover, there is a set of orthonormal vectors $\br_1, \dots, \br_K \in \bbR^K$ such that, for $i \in I_k$,
$$\bru_i =  a_{ik} \br_k, \text{ with } a_{ik} = \sqrt{\frac{\rD_i}{\sum_{j \in I_k} \rD_j}}; \text{ and therefore } \brv_i =  \br_k.$$
\end{proposition}
The proof of Proposition \ref{prop:inf} is in Section \ref{proof:inf}.

\subsection{The case of finite separation}


For a subspace $A$, let ${\rm Proj}_A$ denote the orthogonal projection onto $A$.  Also, let $\| \cdot \|_F$ denote the (matrix) Frobenius norm.  Let $\bu_1, \dots, \bu_N$, and $\bv_1, \dots, \bv_N$, denote the row vectors of $\bU$ and $\bV$ respectively.

\begin{lemma}
\label{lem:2bases}
Let $A$ and $B$ be two linear subspaces in $\bbR^n$ of same dimension $p$.  For every orthonormal basis $\{\ba_1, \dots, \ba_p\}$ of $A$, there is orthonormal basis $\{\bb_1, \dots, \bb_p\}$ of $B$ such that
$$\sum_{j = 1}^p \|\ba_j - \bb_j\|^2 = \|{\rm Proj}_A - {\rm Proj}_B\|_F^2.$$
\end{lemma}
The proof of Lemma \ref{lem:2bases} is in Section \ref{proof:2bases}.

Let $L$ (resp. $\rL$) denote the subspace spanned by the top $K$ eigenvectors of $\bZ$ (resp. $\brZ$).  By Lemma \ref{lem:2bases}, given an orthonormal basis $\bU$ for $L$ (in matrix form), there is an orthonormal basis $\brU$ for $\rL$ such that
$$\|\bU - \brU\|_F^2 = \|{\rm Proj}_L - {\rm Proj}_{\rL}\|_F^2.$$
Note that $\|\bU - \brU\|_F^2 = \sum_{i = 1}^N \|\bu_i - \bru_i\|^2$.  Now, by the triangle inequality,
$$\|\bv_i - \brv_i\| \leq \|\bu_i\| (\|\bu_i\|^{-1} - \|\bru_i\|^{-1}) + \|\bu_i - \bru_i\|  \|\bru_i\|^{-1} \leq 2 \|\bu_i - \bru_i\|  \|\bru_i\|^{-1}.$$
By Proposition \ref{prop:inf} and (A4), $\|\bru_i\|^{-2} \leq \theta N_k \leq \theta N$, so that
$$\sum_{i = 1}^N \|\bv_i - \brv_i\|^2 \leq 2 \theta N \sum_{i = 1}^N \|\bu_i - \bru_i\|^2.$$
By Proposition \ref{prop:inf}, there is a set of orthonormal vectors $\br_1, \dots, \br_K \in \bbR^K$ such that, for $i \in I_k$, $\brv_i = \br_k$, and therefore satisfies 
$$\sum_{k = 1}^K \sum_{i \in I_k} \|\bv_i - \br_k\|^2 \leq 2 \theta N \|{\rm Proj}_L - {\rm Proj}_{\rL}\|_F^2.$$

As in~\cite{spectral_theory}, we use~\cite[Th. 3]{Zwald06}, together with Proposition \ref{prop:inf} to obtain the following bound
$$\|{\rm Proj}_L - {\rm Proj}_{\rL}\|_F \leq 2 \|\bZ - \brZ\|_F / \zeta.$$

We then conclude with the following bound on the amount of perturbation.

\begin{proposition}
\label{prop:perturb}
Assume (A2)-(A3) are satisfied.  Then, $\|\bZ - \brZ\|_F \leq \sqrt{K^2 \nu_1 + K \nu_2^2}$.
\end{proposition}
The proof of Proposition \ref{prop:perturb} is in Section \ref{proof:perturb}.

\section{Proofs of Auxiliary Results}
\label{sec:aux}

\subsection{Proof of Lemma \ref{lem:vol1}}
\label{proof:vol1}

Let $\by \in S$ such that $\|\bx - \by\| \leq \tau$.  First, consider the case $\tau \geq \eps/2$.  The upper bound is obvious.  For the lower bound, consider $\bz = (1 - \lambda) \bx + \lambda \by$ with $\lambda = \eps/(4\tau)$, so that $\|\bz - \bx\| \leq \eps/4$ and $\|\bz - \by\| \leq \tau - \eps/4$, and therefore $B(\bz,\eps/8) \subset B(S,\tau) \cap B(\bx, \eps)$,  by the triangle inequality, with $B(\bz,\eps/8) \asymp \eps^D$.  Next, assume $\tau \leq \eps/2$.  Note that $B(\by, \eps/2) \subset B(\bx, \eps) \subset B(\by, (3/2) \eps)$, so we focus on proving Lemma \ref{lem:vol1} for $\by$ in place of $\bx$.  Let $\by_1, \dots, \by_n$ be a $\tau$-packing of $S \cap B(\by, \eps-\tau)$.  We have $n \asymp (\eps/\tau)^d$; indeed, 
$$S \cap B(\by, \eps/2) \subset \bigcup_j S \cap B(\by_j, \tau) \Rightarrow \eps^d \asymp \vol_d(S \cap B(\by, \eps/2)) \leq \sum_j \vol_d(S \cap B(\by_j, \tau)) \asymp n \tau^d;$$ 
$$S \cap B(\by, \eps) \supset \bigsqcup_j S \cap B(\by_j, \tau/2) \Rightarrow \eps^d \asymp \vol_d(S \cap B(\by, \eps)) \geq \sum_j \vol_d(S \cap B(\by_j, \tau/2))\asymp n \tau^d.$$ 
In a similar fashion, using the triangle inequality,
$$B(S,\tau) \cap B(\by, \eps/2) \subset \bigcup_j B(\by_j, 2\tau) \Rightarrow \vol_D(B(S,\tau) \cap B(\by, \eps/2)) \leq \sum_j \vol_D(B(\by_j, 2\tau)) \asymp n \tau^D;$$ 
$$B(S,\tau) \cap B(\by, \eps) \supset \bigsqcup_j B(\by_j, \tau/2) \Rightarrow \vol_D(B(S,\tau) \cap B(\by, \eps)) \geq \sum_j \vol_D(B(\by_j, \tau/2)) \asymp n \tau^D.$$
Therefore, $\vol_D(B(S,\tau) \cap B(\bx, \eps)) \asymp n \tau^D$, which together with $n \asymp (\eps/\tau)^d$ implies Lemma \ref{lem:vol1}.  This conclude the proof of Lemma \ref{lem:vol1}.

\subsection{Proof of Lemma \ref{lem:1-detect}}
\label{proof:1-detect}

Define $\zeta_j$ as equal to the number of points in $(f_j - \delta/2, f_j + \delta/2)$.  Let $m' = |J_1| = |J_0|$, and note that $m' \sim m/4$.  We put uniform priors on both the null and the alternative.  The resulting likelihoods (with respect to the uniform measure) under the null and alternative are 
$$Z_s = (m')^{-1} (1 - 1/m)^{-N} \cdot \# \{j \in J_s: \zeta_j = 0\}, \ s = 0, 1.$$  
Note that $Z_0$ and $Z_1$ have the same distribution under the uniform measure.  It is enough to show that the Hellinger affinity, here $\expect{\sqrt{Z_0 Z_1}}$, tends to one~\cite[Th. 13.1.2]{MR2135927}; we do that by showing that $\liminf \expect{\sqrt{Z_0 Z_1}} \geq 1$.  Since $\expect{\sqrt{Z_0 Z_1}} \geq \expect{\sqrt{Z_0 Z_1} \cdot {\bf 1}\{Z_0, Z_1 < 2\}}$, by dominated convergence it is enough to show that $Z_1, Z_0 \to 1$ in probability.  For that, we prove that ${\rm var}(Z_0) = o(1)$, which is enough since $\expect{Z_0} = 1$.  From 
$$
{\rm var}({\bf 1}\{\zeta_j = 0\}) \leq (1 - 1/m)^N, \quad
{\rm cov}({\bf 1}\{\zeta_j = 0\}, {\bf 1}\{\zeta_k = 0\}) = (1 - 2/m)^N - (1 - 1/m)^{2N} < 0,
$$
we have
$$
{\rm var}(Z_0) \leq (m')^{-2} (1-1/m)^{-2N} \cdot (m' (1 - 1/m)^N) \prec (1/m) \exp(N/m).
$$
Therefore, ${\rm var}(Z_0) = o(1)$ when $m \geq C^{-1} \log(N)/N$, i.e. $\delta \leq C \log(N)/N$, where $C < 1$.

\subsection{Proof of Lemma \ref{lem:2-detect}}
\label{proof:2-detect}

Define $\zeta_{0, \pi, j}$ as equal to the number of points in $((j-1) A \delta, j A \delta) \times (f_{0, \pi, j} - \delta/2, f_{0, \pi, j} + \delta/2)$, where $f_{0, \pi, j} = 1/4 + \pi_1 + \cdots + \pi_j$; also, let $\zeta_{0,\pi} = \sum_j \zeta_{0,\pi,j}$.  We define $\zeta_{1, \pi, j}$ and $\zeta_{1,\pi}$ similarly, with $1/4$ replaced by $3/4$.  We see $\pi$ as the increments of a nearest-neighbor path in $\mathbb{Z}$.  As such, we put the same prior on the null and the alternative, defined by the random path with low-predictability profile used in~\cite[Sec. 2.2.4]{MR2435454} and inspired from~\cite{MR1640343,MR1634419}.  The resulting likelihoods (with respect to the uniform measure) under the null and alternative are 
$$Z_s = (1-1/(A m))^{-N} \cdot \pr{\zeta_{s,\pi} = 0 | \bx_1, \dots, \bx_N}, \ s = 0, 1.$$
Following the proof of Lemma \ref{lem:1-detect}, we show that ${\rm var}(Z_0) = o(1)$.  We have 
$$
\expect{Z_0^2} = (1-1/(A m))^{-2N} \expect{\pr{\zeta_{\pi} = \zeta_{\pi'} = 0 | \pi, \pi'}},
$$
where $\pi, \pi'$ are two independent copies distributed according to the prescribed prior on paths.  Let $T = |\pi \cap \pi'|$; because,
$$
\pr{\zeta_{\pi} = \zeta_{\pi'} = 0 | \pi, \pi'} = (1 - |\pi \cup \pi'|/(A m^2))^N, \text{ with } |\pi \cup \pi'| = 2 m - T,
$$
we have
$$
\expect{Z_0^2} = (1-1/(A m))^{-2N} \expect{(1 - (2m - T)/(A m^2))^N} \leq e^{2N/(A m)^2} \expect{ e^{T \cdot N/(A m^2)} }.
$$
Following the arguments in~\cite[Sec. 2.2.4]{MR2435454}, we get 
$$
\expect{ e^{T \cdot N/(Am^2)} } \leq \exp( C \log(m)^2 \log\log(m) N/m^2).
$$
Therefore, $\expect{Z_0^2} \to 1$ when $\log(m)^2 \log\log(m) N/m^2 = o(1)$, which happens when 
$$\delta \ll N^{-1/2} \log(N)^{-1} \log\log(N)^{-1/2}.$$

\subsection{Proof of Proposition \ref{prop:inf}}
\label{proof:inf}

Without loss of generality, we assume that the points are indexed so that $I_k = \{N_1 + \cdots + N_{k-1} + 1, \dots, N_1 + \cdots + N_{k-1} + N_k\}$.  With this ordering, $\brW$ is block diagonal, $\brW = \diag\{\bW_1, \dots, \bW_K\}$, and so is $\brZ = \diag\{\brZ_1, \dots, \brZ_K\}$, with $\brZ_k = \brD_k^{-1/2} \bW_k \brD_k^{-1/2}$ and $\brD_k = \diag\{\bW_k \cdot {\bf 1}\}$.

Each $\brZ_k$ has the same spectrum as $\brD_k^{-1} \bW_k$, which is the transition matrix of the Markov chain associated with the graph with affinity matrix $\bW_k$.  Indeed, $\bv$ is an eigenvector for $\brD_k^{-1} \bW_k$ if, and only if, $\brD_k^{1/2} \bv$ is an eigenvector for $\brZ_k$.  Therefore, the largest eigenvalue of $\brZ_k$ is 1, and because of (A1), that eigenvalue is simple.  Hence, the largest eigenvalue of $\brZ$ is 1 and $\brZ$ has that eigenvalue with multiplicity $K$, which is the number of blocks.  Moreover, the eigengap for $\brZ$ is the minimum among the eigengaps of the $\brZ_k, k = 1, \dots, K$, which are all bounded below by $\zeta$ when (A1) holds.  

Concerning eigenvectors, $\ba_k = \brD_k^{1/2} {\bf 1} / \|\brD_k^{1/2}\|_F$ is a top (normalized) eigenvector for $\brZ_k$, since $\brD_k^{-1} \bW_k \cdot {\bf 1} = {\bf 1}$.  Let $\bb_k = ({\bf 0}, \ba_k, {\bf 0})^T \in \bbR^N$, where the first (resp. second) ${\bf 0}$ is of size $N_1 + \cdots + N_{k-1}$ (resp. $N_{k+1} + \cdots + N_{K}$).  Any system of $K$ orthogonal eigenvectors of $\brZ$ for the eigenvalue 1 is a rotation of $\bB = [\bb_1, \dots, \bb_K]$.  Therefore, there is an orthogonal matrix $\bR = [\br_1, \dots, \br_K]$ such that $\brU = \bB \bR^T$.  That is, for $i \in I_k$, the row vectors of $\brU$ satisfy $\bru_i = a_{ik} \br_k$.  After normalization, we get $\brv_i = \br_k$ for $i \in I_k$.

\subsection{Proof of Lemma \ref{lem:2bases}}
\label{proof:2bases}

Let $\theta_1, \dots, \theta_p$ denote the principal angles between these two subspaces~\cite{MR1417720}.  It is well-known that 
$$\|{\rm Proj}_A - {\rm Proj}_B\|_F^2 = 2 \sum_{j = 1}^p \sin^2 \theta_j.$$
(The quantity $\|{\rm Proj}_A - {\rm Proj}_B\|_F$ is called in~\cite{305358} the {\em projection F-norm distance} between the subspaces $A$ and $B$.)
On the other hand, by definition of the principal angles, there are orthonormal bases, $\{\ba_1^0, \dots, \ba_p^0\}$ for $A$ and $\{\bb_1^0, \dots, \bb_p^0\}$ for $B$, such $\cos \theta_j = (\ba_j^0)^T \bb_j^0$.  Therefore,
$$\sum_{j = 1}^p \|\ba_j^0 - \bb_j^0\|^2 = 2 \sum_{j = 1}^p \sin^2 \theta_j.$$
We then simply choose an orthogonal matrix $\bR$ such that $\ba_j = \bR \ba_j^0$ and define $\bb_j = \bR \bb_j^0$.

\subsection{Proof of Proposition \ref{prop:perturb}}
\label{proof:perturb}

This comes from directly summing over blocks.  Indeed,
\begin{equation} \label{eq:Z-diff}
\|\bZ - \brZ\|_F^2 = \sum_{k} \sum_{i, j \in I_k} (Z_{i j} - \rZ_{i j})^2 + \sum_{k \neq \ell} \sum_{i \in I_k} \sum_{j \in I_\ell} Z_{i j}^2.
\end{equation}

Note that for all $i \in I_k$, $D_i - \rD_i = \sum_{j \notin I_k} W_{ij}$ so that $D_i - \rD_i \geq 0$ and by (A3) we have
\begin{equation} \label{eq:Dratio}
\frac{D_i}{\rD_i} \leq 1 + \nu_2 C_k^{-1/2}, \quad C_k := \sum_{s, t \in I_k} \frac{W_{st}^2}{\rD_{s} \rD_{t}}.
\end{equation}

For $i, j \in I_k$, using (\ref{eq:Dratio}) we get
$$(Z_{i j} - \rZ_{i j})^2 =  \frac{W_{ij}^2}{\rD_i \rD_j} \left( \sqrt{\frac{D_i}{\rD_i} \frac{D_j}{\rD_j}} - 1 \right)^2 \leq \frac{W_{ij}^2}{\rD_i \rD_j} \cdot \nu_2^2 C_k^{-1}.$$
Hence, the first term on the right hand side of (\ref{eq:Z-diff}) is bounded from above by $K \nu_2^2$.

For $i \in I_k$ and $j \in I_\ell$, $Z_{i j}^2 = W_{ij}^2/(D_i D_j) \leq W_{ij}^2/(\rD_i \rD_j)$, since $\rD_i \leq D_i$ for all $i$.  Hence, by (A2), the second term on the right hand side of (\ref{eq:Z-diff}) is bounded from above by $K^2 \nu_1$.

\section*{Acknowledgment}

The author got interested in spectral clustering out of discussions with Gilad Lerman, who shared early versions of his manuscripts, joint with Guangliang Chen, on clustering and detecting affine subspaces using spectral methods.  The author is hence grateful to both of them.  The author is indebted to Andrew Ng for providing the arguments underlying the results presented in his paper~\cite{Ng02}.  In addition, the author would like to thank James Bunch, Lawrence Cayton and Ofer Zeitouni for providing references and for entertaining stimulating discussions.  
Some anonymous referees from NIPS (where a short version of this paper was originally sent), provided references~\cite{1519716,RePEc:eee:stapro:v:35:y:1997:i:1:p:33-42} and the author is grateful for that.  This work was partially supported by NSF grant DMS-06-03890 and ONR grant N00014-09-1-0258.

\bibliographystyle{abbrv}
\bibliography{refs}

\end{document}